\documentclass[10pt]{iopart}
\usepackage{xcolor}
\usepackage{amsmath}
\usepackage{graphics} 
\usepackage{epsfig} 
\usepackage{caption}
\usepackage{subcaption}
\usepackage{textcomp}
\usepackage[inline]{enumitem}
\usepackage{makecell}
\usepackage{nicematrix}
\usepackage[flushleft]{threeparttable}

\begin{document}

\title{Neuromorphic computing for attitude estimation onboard quadrotors}


\author{Stein Stroobants, Julien Dupeyroux and Guido C.H.E. de Croon}

\address{Micro Air Vehicle Lab, Faculty of Aerospace Engineering, Delft University of Technology, The Netherlands}
\ead{s.stroobants@tudelft.nl}
\vspace{12pt}
\begin{indented}
\item[]March 2022
\end{indented}

\begin{abstract}
Compelling evidence has been given for the high energy efficiency and update rates of neuromorphic processors, with performance beyond what standard Von Neumann architectures can achieve. Such promising features could be advantageous in critical embedded systems, especially in robotics. To date, the constraints inherent in robots (e.g., size and weight, battery autonomy, available sensors, computing resources, processing time, etc.), and particularly in aerial vehicles, severely hamper the performance of fully-autonomous on-board control, including sensor processing and state estimation. In this work, we propose a spiking neural network (SNN) capable of estimating the pitch and roll angles of a quadrotor in highly dynamic movements from 6-degree of freedom Inertial Measurement Unit (IMU) data. With only 150 neurons and a limited training dataset obtained using a quadrotor in a real world setup, the network shows competitive results as compared to state-of-the-art, non-neuromorphic attitude estimators. The proposed architecture was successfully tested on the Loihi neuromorphic processor on-board a quadrotor to estimate the attitude when flying. Our results show the robustness of neuromorphic attitude estimation and pave the way towards energy-efficient, fully autonomous control of quadrotors with dedicated neuromorphic computing systems. 
\end{abstract}


%
\vspace{2pc}


\section{Introduction}
Over the last two decades, efforts have been made to combine the fields of Artificial Intelligence (AI) and Robotics with outstanding results~\cite{Rajan2017}. 
Algorithms making use of AI techniques, such as deep neural networks, have been proposed to achieve state estimation~\cite{Ha2018}, object manipulation~\cite{Agrawal2016}, localization~\cite{Giusti2016} and control~\cite{Loquercio2021, Poznyak2001, Lillicrap2016}. For instance in~\cite{Hwangbo2017}, a quadrotor learns to fly by applying Reinforcement Learning (RL) to a densely connected neural network. 
Aerial vehicles are critical embedded systems, the constraints of which (e.g., size and weight, battery autonomy, sensors, computing resources) hamper the development and the performance of fully autonomous and robust on-board control. 
In particular, Micro Air Vehicles (MAVs) are a class of aerial vehicles that could strongly benefit from AI-powered solutions to handle their highly non-linear dynamics, and allow for online adaptations to unpredictable changes occurring in the real world (e.g., gusts of wind, sensor damage, communication failure, etc.). 
Many of the complex tasks future MAVs will have to perform will be powered by AI. One can think of deep neural networks that have to estimate optical flow~\cite{Ilg2017} and recognize objects~\cite{Redmon2016}, which can be used for vision-based autonomous navigation in urban areas to deliver packages.


However, MAVs are now locked out from using large-scale neural networks because of the great amount of energy required, as well as disproportionate need for computing resources that only Graphics Processing Units (GPUs) can offer. Additionally, standard Von Neumann architectures suffer from a relatively high latency that restrict their use in extreme conditions such as drone racing and aggressive flight maneuvers.


Alternatively to traditional Artificial Neural Networks (ANNs) where the information is processed in a synchronous manner, Spiking Neural Networks (SNNs) could represent the choice solution to bridge the gap between AI and resource-restricted Robotics. In contrast to ANNs, SNNs encode the information not by the intensity of the signal, but by a series of binary events, also called spikes, and the relative time between them. Inspired by their biological counterpart, spiking neurons accumulate incoming synaptic currents over time and fire whenever their membrane potential exceeds a certain threshold. While a wide range of neuron models have been proposed, the most commonly used are the Integrate-and-Fire (IF) and Leaky-Integrate-and-Fire (LIF)~\cite{Kasabov2019}. 
The simpler binary signals and sparse firing of SNNs hold the promise of orders of magnitude more energy-efficient processing than ANNs~\cite{Cao2015}.

The move towards SNNs requires a complete shift in the coding and processing of information that is not optimized for Von Neumann architectures. In this regard, neuromorphic sensors and processors have been designed, arousing the enthusiasm of roboticists to embed these new technologies onboard robots. Examples of neuromorphic processors include HICANN~\cite{Schemmel2010}, NeuroGrid~\cite{Benjamin2014}, IBM's TrueNorth~\cite{Merolla2014}, APT's SpiNNaker~\cite{Furber2014} which is part of the Human Brain Project ~\cite{Calimera2013}, and Intel's Loihi~\cite{Davies2018}. In terms of neuromorphic sensing, most efforts have been put to develop the neuromorphic equivalent to standard CMOS cameras, namely, event-based cameras~\cite{Gallego2022}. Neuromorphic tactile sensors have also been proposed~\cite{Caviglia2016}.

Tackling computationally expensive vision or navigation tasks with neuromorphic sensing and processing will bring large energy and speed benefits. However, to optimally reap the benefits of neuromorphic algorithms and hardware, an end-to-end, fully neuromorphic solution needs to be designed. Then a single neuromorphic processor could suffice. In the literature, some of the steps towards such a fully neuromorphic pipeline have been demonstrated. In~\cite{Vitale2021}, the authors introduced a neuromorphic PD controller that outperforms state-of-the-art controllers in high-speed control thanks to the synergy between the high update rates of the neuromorphic chip and the event-camera. Also, autonomous thrust control of a flying quadrotor was achieved by an SNN evolved in simulation, using optic flow to enact a constant-divergence landing~\cite{dupeyroux2021neuromorphic}. 
In~\cite{Rueckert2016}, the authors show that SNNs are capable of solving planning tasks, such as avoiding obstacles with a robotic arm. 

This fully neuromorphic pipeline will also have to include "lower-level" tasks, but these are not well studied until now. For example, autonomous control onboard MAVs requires an accurate estimate of the states, such as attitude and global lateral position and velocity, by combining sensor measurements. In this article, we propose a neuromorphic solution for onboard attitude estimation of a MAV using data from an inertial measurement unit (IMU), combining data from a 3-axes accelerometer and a 3-axes gyroscope. We compare the network to a similarly sized and trained traditional Recurrent Neural Network (RNN) and commonly used filters specific to this task. The proposed SNN is trained from limited data obtained with a real MAV and can be employed as part of an autonomous neuromorphic flight-controller pipeline for an MAV. Closest to our work is the study in~\cite{Weber2020}, which trains a traditional ANN with recurrency for attitude estimation. Specifically, in~\cite{Weber2020} it is shown that a 2-layer Recurrent Neural Network (RNN) can be trained to perform this task with outstanding results on pre-gathered datasets. This network, however, is not spiking and has not been applied in the control loop of MAVs in flight.


MAVs are usually equipped with an IMU and use the combination of the angular velocities and the linear accelerations to estimate the current attitude. Angular velocities show the rate of change, but induce integration errors over longer time-windows while the linear acceleration shows the gravity vector over longer stretches of time. The output pitch and roll estimates are necessary for the drone to be able to control the position in the x-y plane by performing attitude control. 

Our contributions are threefold. First, we propose an SNN architecture to perform state estimation for dynamic systems such as quadrotors with limited data obtained with a physical quadrotor and ground-truth. Then, we demonstrate that the proposed neuromorphic state estimator exhibits competitive performance when compared to widely used, non-neuromorphic solutions (i.e., Madgwick filter, Mahony filter, and complementary filter) and to a traditional RNN. Lastly, we successfully test our solution onboard a quadrotor equipped with the Loihi neuromorphic chip, thus paving the way for a fully neuromorphic control-loop for quadrotors.

\section{Methods}
\subsection{Spiking network definition}
\begin{figure}[t]
    \centering
    \includegraphics[width=1\linewidth]{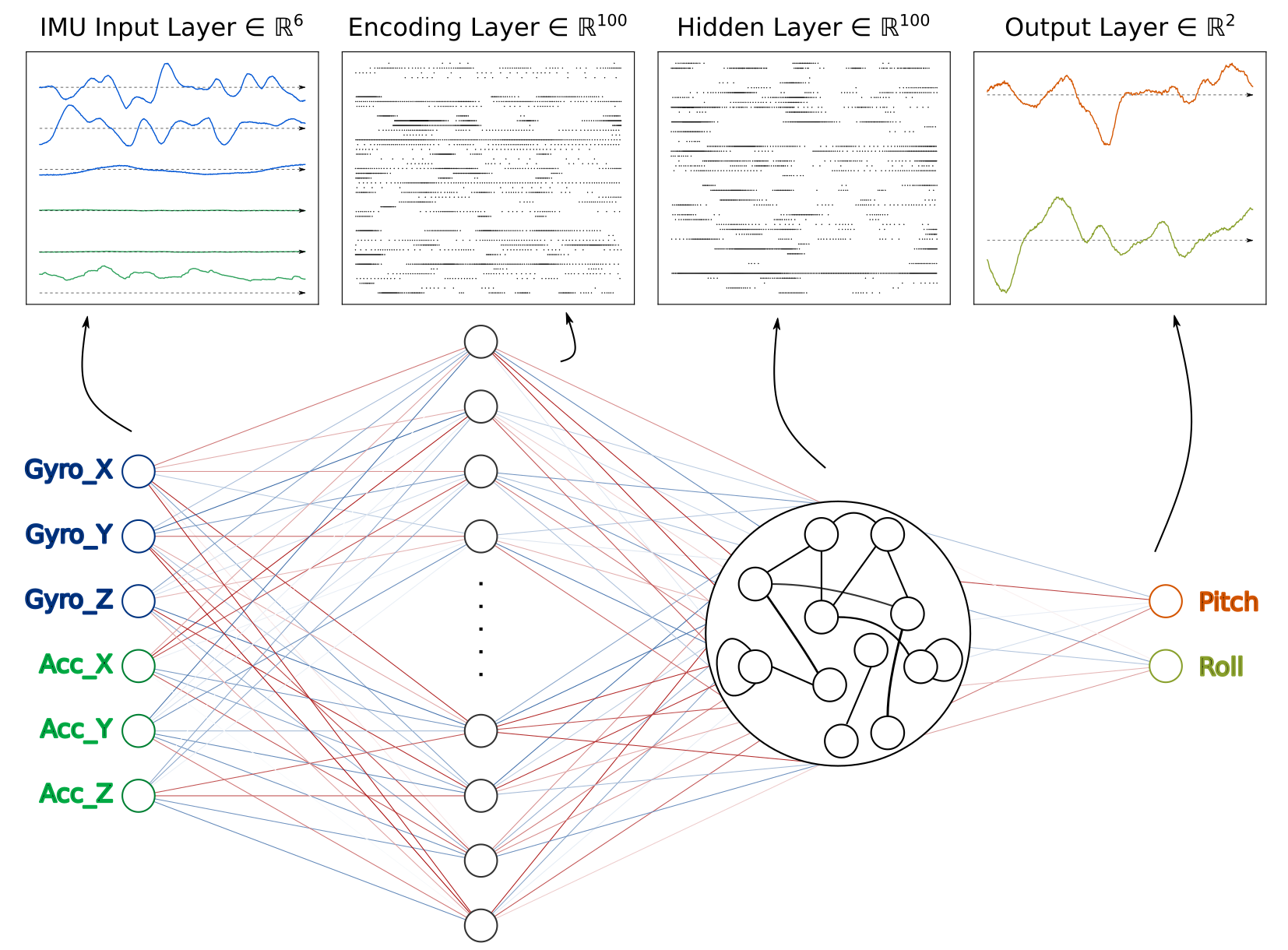}
    \caption{Topology of the attitude estimation network Att-SNN showing signals transported between layers. In the encoding layer, the normalized IMU data is transformed into spikes. The next layer is a fully connected recurrent layer that sends spikes to the Leaky-Integrator (LI) output layer, converting spikes back to attitude estimates.}
    \label{fig:topology}
\end{figure}

This section introduces the attitude estimation spiking neural network, called Att-SNN. The different components that make up the network are shown in Figure~\ref{fig:topology}. 

\subsubsection{Spiking neuron model}


In this work, we use the Leaky-Integrate-and-Fire (LIF) neuron as the core of our SNN. Widely used in the literature, the LIF model is available in most SNN simulators a neuromorphic hardware, including the  Loihi used for this study~\cite{Davies2018}. The discrete-time difference equations governing the LIF neuron are given as follows:

\begin{align}
    \label{eq:LIF}
    \upsilon_i(t+1) &= \tau^{\text{mem}}_ i\upsilon_i(t) + i_i(t) \\
    i_i(t+1) &= \tau^{\text{syn}}_i i_i(t) + \sum w_{ij} s_j(t)
\end{align}

\noindent where $\upsilon_i(t)$ is the membrane potential at time $t$, $\tau^{\text{mem}}_i \in [0,1]$ and $\tau^{\text{syn}}_i \in [0,1]$ the membrane and synaptic time constants, $i(t)$ the synaptic current at time $t$, $w_{ij}$ the synaptic weight between neurons $i$ and $j$, and $s_j$ a binary value representing either a spike or no spike coming from the pre-synaptic neuron $j$. To determine whether a neuron emits a spike, the membrane potential is reduced with the neurons firing threshold $\theta^{\text{thr}}_i$ and passed through the Heaviside step-function to determine the output of the neuron:
\begin{equation}
    s_i(t) = H(\upsilon_i(t) - \theta^\text{thr}_i)=\begin{cases} 0, & \upsilon_i(t)  - \theta^\text{thr}_i \leq 0 \\ 1, & \upsilon_i(t)  - \theta^\text{thr}_i > 0 \end{cases}
    \label{eq:heaviside}
\end{equation}

When the Heaviside-function resolves to 1 and the neuron emits a spike, the membrane potential $\upsilon_i(t)$ is reset to zero. 

\subsubsection{Data encoding}

Standard, off-the-shelf IMUs are not neuromorphic; the output data is formatted as floating point values and streamed synchronously. As a result, the measured angular rates and linear accelerations must be translated into spikes so that they can be processed by the SNN. 

Data encoding is a complex task. Spike coding algorithms can be divided into three categories. Population coding uses set of distinct neurons to encode (or decode) the signal by emitting a spike whenever the input signal falls within the range distribution of one (or several) neuron. It has been successfully applied in~\cite{Stagsted2020, dupeyroux2021neuromorphic}. Temporal coding algorithms encode the information with high timing precision, by emitting a spike whenever the variation of the input signal exceeds a threshold. Rate coding, which was used in~\cite{Rueckert2016}, encodes the information into the the firing frequency of a population of neurons. 


In this work, the floating point values returned by the 6-DOF IMU are encoded into spikes by means of a spiking layer densely connected to the 6 outputs of the IMU sensor (Fig.~\ref{fig:topology}). The synaptic weights, as well as the dynamics of the neurons in this encoding layer were learned offline. Both the time-constants were trained for each neuron seperately, resulting in $2N$ parameters, with $N$ the number of neurons in the encoding layer. Before passing the IMU data to the network, a normalization process is applied to ensure faster learning and convergence during training. This normalization is achieved through min-max, mapping the input data to a range of $[-1,1]$, according to the formula hereafter. The values of $x_\text{min}$ and $x_\text{max}$ were chosen empirically to match the full range of possible inputs.

\begin{equation}
    x_\text{norm}(t) = 2 \cdot \frac{x(t) - x_\text{min}}{x_\text{max} - x_\text{min}} - 1
\end{equation}

\subsubsection{Decoding the global attitude}

Decoding of the spiking activity to an estimate of the attitude in radians is performed via a non-spiking decoding layer (Fig.~\ref{fig:topology}), composed of two Leaky-Integrate (LI) neurons. In this case, the membrane potential $v_i(t)$ in Equation \ref{eq:LIF} of the neuron is directly used as the output of the layer, providing a stateful decimal value representing the current pitch or roll attitude. In contrast to the input-data, no further normalization is required for the attitude values. The pitch and roll angles in the training data, expressed in radians, are already distributed around a mean of zero with standard deviation of $0.25$. Theoretically, the limits go from $-\pi$ to $\pi$, but since we are targeting a quadrotor in a normal flight regime around hover, these values will not be reached.  Lastly, since the pitch and roll axes of the MAV are symmetric (this is true in the case of a standard, symmetric quadrotor), the neuron parameters of the decoding layer are set equal during training. This results in training two parameters for this layer. 


\subsection{Training}

\subsubsection{Training setup}

\begin{figure}[t]
     \centering
     \begin{subfigure}[b]{0.48\textwidth}
         \centering
        \includegraphics[width=\textwidth]{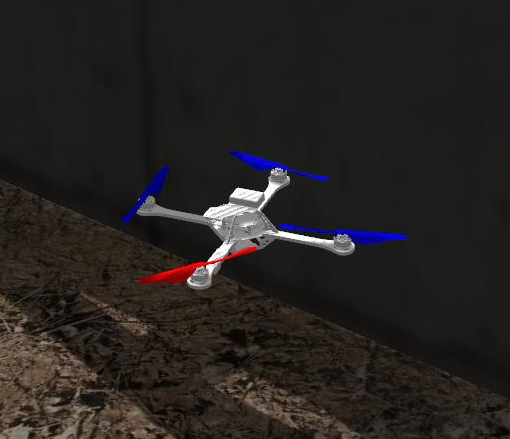}
        \caption{AscTec Hummingbird flying in open-source RotorS quadrotor simulator \cite{Furrer2016}, used to gather supplementary simulation data for training.}
        \label{fig:sim_drone}
     \end{subfigure}
     \hfill
     \begin{subfigure}[b]{0.48\textwidth}
             \centering
            \includegraphics[width=\textwidth]{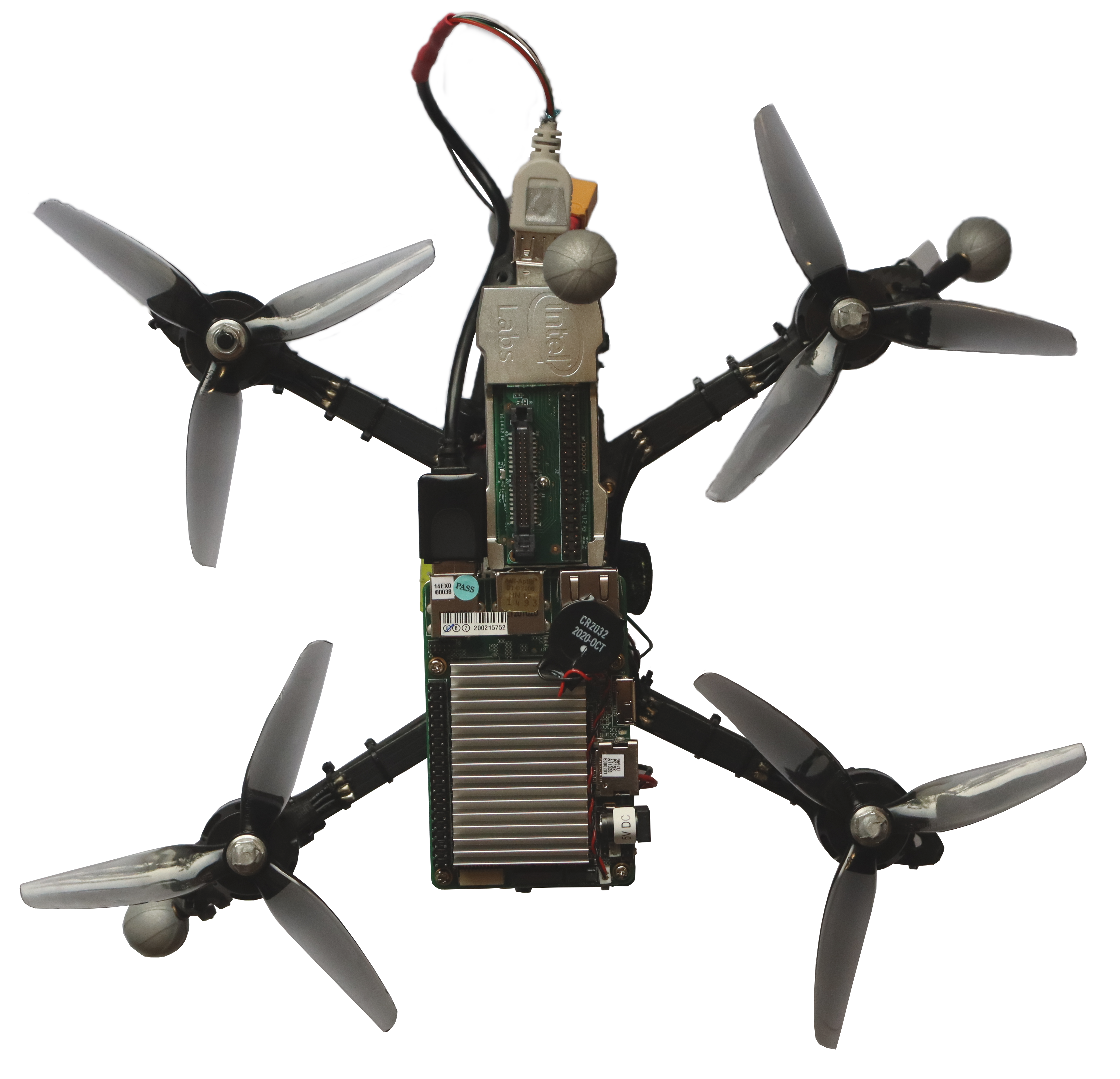}
            \caption{Quadrotor used in our research to gather real-world data and for the evaluation of our SNN on the neuromorphic processor Loihi). }
            \label{fig:real_drone}
     \end{subfigure}
        \caption{The quadrotors used for the simulation (RotorS Gazebo) and real-world tests (data acquisition and tests on the Loihi neuromorphic chip).}
        \label{fig:drones}
\end{figure}

\begin{figure}[t]
    \centering
    \includegraphics[width=0.8\linewidth]{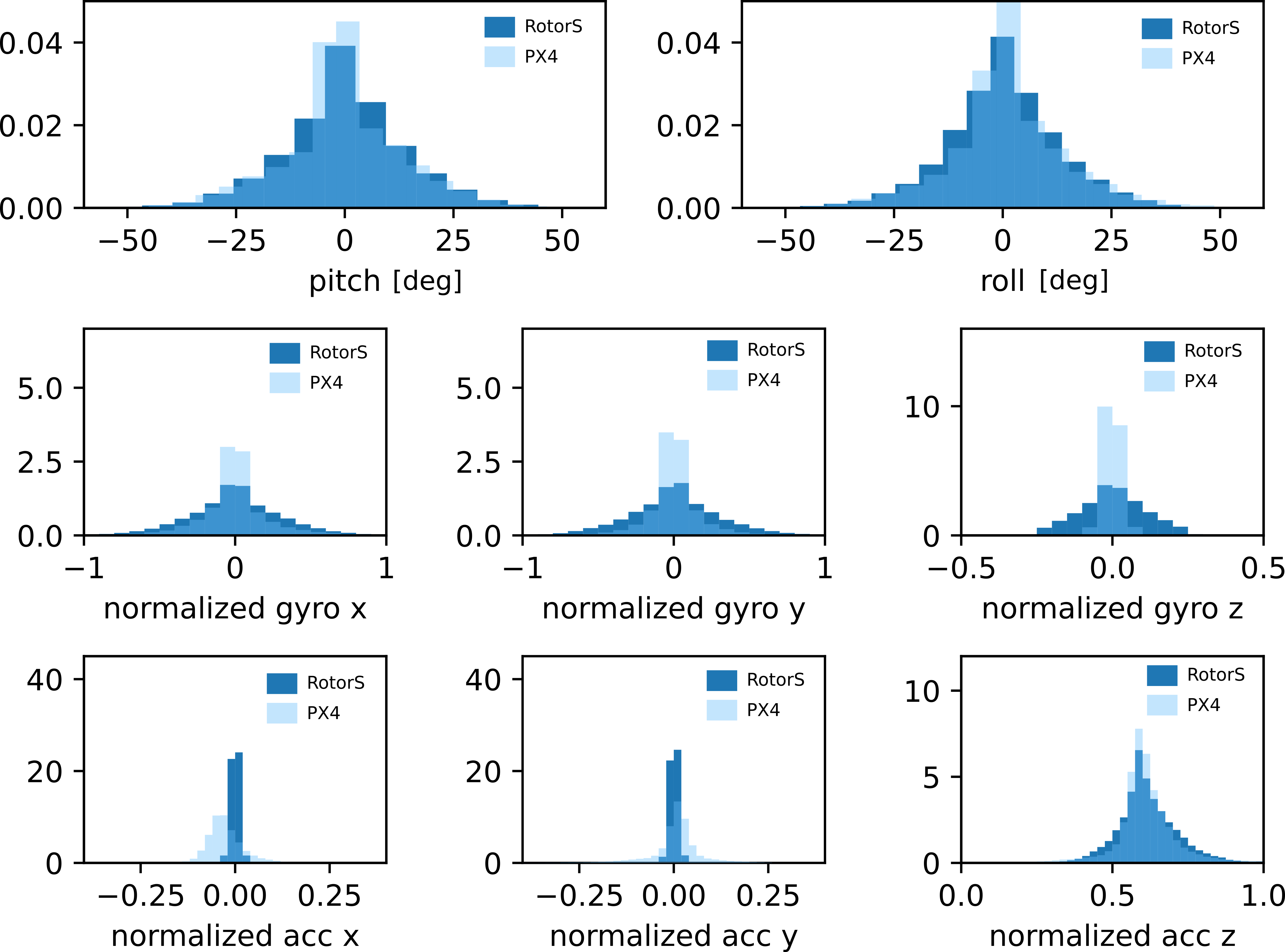}
    \caption{Distribution of the input-output data used for training the network. The inputs (lower six histograms) show sensor-readings after applying normalization. }
    \label{fig:data_analysis}
\end{figure}

The proposed Att-SNN was trained both in simulation, using datasets created with the RotorS simulator~\cite{Furrer2016} (Fig.~\ref{fig:sim_drone}), and with data collected with a quadrotor flying in the real world\footnote{The dataset will be made available online upon publication} (Fig.~\ref{fig:real_drone}). The quadrotor is equipped with a Pixhawk 4 Mini flight controller, which combines the measurement of two separate IMU's for redundancy. The IMU data are logged at 200Hz. For these experiments, the ground truth was provided by a multi-camera motion capture system (OptiTrack). This system provides sub-degree accuracy attitude estimates at the same rate as the combined IMU measurements from the quadrotor. An overview of the distribution of the IMU and OptiTrack data collected both in simulation and in the real world is provided in Figure~\ref{fig:data_analysis}. 
In total, 35 datasets of 100 seconds were gathered in simulation and 11 datasets of 100 seconds in our real-world tests. This amounts to a total of 77 minutes of flight time. 
Both the simulation and real-world data sets have been gathered in a way as to represent roll and pitch angles between hover and swift flight, reaching relatively large angles up to 45 degrees. This covers a normal flight regime. An important challenge in IMU-based attitude estimation is that the gyros and accelerometers have biases that can change over time. 
For instance, accelerometers usually suffer from a turn-on bias which shows as a constant offset of the measured acceleration. This bias is especially visible in the $x$-axis of the PX4 accelerometer. 
It is also worth noting that there is also a difference in the range of values in the $gyro_z$, which shows that the quadrotor in simulation was rotating more around the z-axis than in the real-world experiments. Since this rotation is only affecting the yaw angles directly, the influence on the pitch and roll estimates is limited.
The datasets were split up in $70\%$ train and $20\%$ validation, and $10\%$ test. 
The training datasets were split into sequences of 10 seconds containing 2000 time steps. Furthermore, the simulated data was augmented by adjusting the accelerometer turn-on bias and both the accelerometer and gyroscope noise densities to reduce the reality gap. 
The noise characteristics were determined using the steady-state error of the real world data set. \\
The Att-SNN was implemented using the Norse~\cite{Norse2021} python library, based on PyTorch. The Adam optimizer~\cite{Kingma2014} was used with a learning rate of 0.005, combined with the k-step ahead, 1 step back Lookahead~\cite{Zhang2019} optimizer to speed up learning. For the Lookahead optimizer, the value of $\alpha$ was set at 0.5 and k at 6.
All code to reproduce the results can be found in~\footnote{All code will be made available online upon publication}.

Since the Heaviside function used in the neuron dynamics (Eq.~\ref{eq:LIF}) is non-differentiable, a surrogate-gradient (SG) was chosen to enable backpropagation through time (BPTT). A summary of SGs can be found in~\cite{Neftci2019}. In this study, we used the SuperSpike~\cite{Zenke2018} with a width of 20 as it is a suitable option for supervised BPTT and is robust to changes in the input paradigm as is the case with our current-based input~\cite{Zenke2021}.

The error between the output and the target was characterized by the Mean-Squared-Error (MSE) loss function where both the pitch- and roll-error were weighted evenly. 
The MSE for a sequence $S$ is calculated as in Equation \ref{eq:mse}, with $\hat{\theta}_k$ and $\theta^{\text{gt}}_k$ the estimated and ground-truth pitch angles at timestep $k$, and  $\hat{\phi}_k$ and $\phi^{\text{gt}}_k$ the estimated and ground-truth roll angles.

\begin{equation}
    \label{eq:mse}
    \text{MSE}_S = \sum_{k=0}^{N=\text{len}(S)} \frac{(\hat{\theta}_k - \theta_k^{\text{gt}})^2 + (\hat{\phi}_k - \phi_{k}^{\text{gt}})^2}{2}
\end{equation}

The Att-SNN is trained with 100 neurons in the encoding layer and 100 in the hidden layer. The thresholds were fixed at $0.5$ for all neurons. This results in $600$ weights and $200$ neuron parameters for the encoding layer, $10~000 + 10~000$ weights and $200$ neuron parameters for the hidden layer and $200$ weights and $2$ neuron parameters for the output layer.  In total, this adds up to training of 20~800 weights and 402 neuron parameters. Every epoch, 15 batches of 40 sequences are randomly selected from all training data (including simulated and real-world data) for a training iteration, and afterwards the error is calculated for all validation sets. The training is stopped by a criterion based on a moving average of the error on the validation dataset over the last 20 epochs. If the moving-average of the error is higher than $110$\% of the lowest average validation loss so far, training is aborted. Besides, if the lowest validation loss so far did not change for at least 50 epochs, training is also aborted.

\subsection{Implementation on neuromorphic hardware}


To fully demonstrate the potential of SNNs for state estimation in MAV applications, the proposed Att-SNN architecture has been implemented on Intel's Loihi processor~\cite{Davies2018}. The constraints of SNN design imposed by the Loihi requires to adapt the SNN parameters, such as the synaptic weights and the neurons' parameters. A naive solution to this would be to quantize the parameters after training, but this would inevitably result in a loss in accuracy, and in the long run could affect the overall performance of the state estimation (and control). Alternatively, we have included the quantization in the training process by replacing the full-resolution weights with their quantized equivalents before the forward pass while propagating the gradients. By doing so, we ensure that the network converges to a solution that is fully compatible with the specific features of the Loihi chip. The quantization function is defined by:
\begin{equation}
    p_q = \text{round}(p / \Delta q) \Delta q, \quad p_q \in [q_\text{min}, q_\text{max}]
\end{equation}

\noindent with $p_q$ the quantized version of parameter $p$, $\Delta q$ the quantization step-size, $\text{round}( . )$ rounding of a floating point value to the closest integer and $[q_\text{min}, q_\text{max}]$ the quantization range. The threshold $\theta_\text{thr}$ was fixed during training, while all other parameters were not. The quantization ranges and step sizes used in this study are:
\begin{enumerate}
    \item[-] synaptic weights $w_{ij}$, range: $[-1, 1 - \frac{1}{256}]$, step size: $\frac{2}{256}$
    \item[-] synaptic decay $\tau_{\text{syn},i}$, range: $[0, 1]$, step size: $\frac{1}{4096}$
    \item[-] membrane decay $\tau_{\text{mem},i}$, range $[0, 1]$, step size: $\frac{1}{4096}$
\end{enumerate}
During implementation on the neuromorphic chip, the parameters can be multiplied by the quantization range, resulting in integers compatible with the constraints set by the neuromorphic processor.


\subsection{Models for comparison}

The performance of the Att-SNN neuromorphic state estimator is compared to the most widely-used non-neuromorphic algorithms: (i) the Madgwick filter~\cite{Madgwick2011}, (ii) the Mahony filter~\cite{Mahony2008}, the Extended Kalman filter (EKF)~\cite{kalman1960new, ribeiro2004kalman}, and (iv) the complementary filter~\cite{Gui2015}. These filters all adequately estimate the pitch and roll angles of a quadrotor in flight. Since only data from a 6-DOF IMU is used, these two are the only observable states out of a full non-linear model of a quadrotor. 
For fairness of comparison, we implement a minimalistic EKF that uses the IMU together with a quaternion-based random walk model to estimate pitch and roll. Typically, EKFs are more extensive, as they can estimate more states such as velocity or position, integrate other sensors like GPS, and can employ more detailed motion models using known control inputs. However, such extensions are out of the scope of the current study.
These filters all require some fine-tuning of their parameters with respect to the dynamic motions captured in the datasets, the noise level measured in the IMU, and a prior knowledge of the initial state. Tuning these parameters by hand is quite common, but this may have resulted in unfair comparison with a neural network that is automatically and thoroughly trained with dedicated algorithms on the training dataset consisting of simulation and PX4 data. As a result, the parameters of all filters were automatically determined for the same datasets using the modified Particle Swarm Optimization (PSO) algorithm as described in~\cite{Shi1998}. 
The optimization values were chosen as: $w = 0.8$, $c_1 = 0.15$ and $c_2 = 0.05$ and a total of 100 particles was used. The cost function was defined as the MSE per timestamp, averaged over all training sequences plus a high cost of $10$ for parameters below 0 or above 1 to constrain the parameter to the corresponding ranges. 
Using the PSO, the optimal parameters were estimated for both the simulation and PX4 datasets, assuming that the filters had no knowledge of the initial angle. The impact of this assumption will be further analyzed in Section~\ref{subsec:perf_analysis}. \\
Additionally, we propose a traditional recurrent neural network, called Att-RNN which is composed of gated recurrent units (GRUs). This network has a very similar structure to our proposed Att-SNN, allowing to evaluate the difference in performance caused by the introduction of the LIF model's neural dynamics and spiking.

\subsubsection{The complementary filter}
The complementary filter is a widely used filter for attitude estimation due to its simplicity. 
The output of the filter is a weighted average between the angle $\theta^\text{acc}_k$ measured by the gravity vector measured with the accelerometer (when the assumption of weak-acceleration relative to gravity holds), and the previous estimated angle $\hat{\theta}_{k-1}$ propagated with the angular velocity $\omega^\text{gyr}_k$ measured by the gyroscopes. 
For a single axis this is defined as follows:
\begin{equation}
     \hat{\theta}_k = \gamma( \hat{\theta}_{k-1} + \omega^\text{gyr}_k \Delta t ) + (1-\gamma) \theta^\text{acc}_k
\end{equation}
Where $\gamma$ is the weighting factor, balancing between the accelerometer that provides a solid baseline on a long time horizon and the gyroscope that is accurate for updating the angle on short time-scale, but suffers from integration errors on a longer scale.

\subsubsection{The Mahony filter}
Gyroscope measurements from low-cost IMUs can be biased, and the Mahony filter is an extension of the complementary filter that counters this~\cite{Mahony2008}.
By adding an integral term $\hat{b}_k$ to the filter, based on the error $e_k$ between the angle measured by the accelerometer and the predicted angle, the gyroscope biases can be effectively canceled without increasing the computationally cost too much.
The filter, in discrete quaternion form, can be written as:
\begin{align}
    \hat{q}_k &= \hat{q}_{k-1} + (\frac{1}{2} \hat{q}_{k-1} \otimes \mathbf{p}\{w^\text{gyr}_k - \hat{b}_k + k_P e_k\}) \Delta t \\
    \hat{b}_k &= \hat{b}_{k-1} - k_I e_k \Delta t
\end{align}
Where $\mathbf{p}\{x\} = [0\ x]^T$ is a pure quaternion that relates to the rotation velocity of the attitude and $p \, \otimes \, q$ is the Hamilton product between quaternion $p$ and $q$. 
This algorithm can be optimized for certain motions or sensors by adjusting the proportional ($k_P$) and integral ($k_I$) gain.

\subsubsection{The Madgwick filter}
Madgwick~\cite{Madgwick2011} defines attitude estimation as a minimization problem, solved with a gradient descent algorithm.
While the gyroscope is still used for integrating the angle, the error with respect to measured gravity is represented as a cost function $f(q)$ describing the difference between the angle measured with the accelerometer and the gravity vector rotated to the body-frame. 
This cost-function is minimized by taking a single gradient descent step.
This results in the following update step in discrete quaternion form:
\begin{equation}
    \hat{q}_k = \hat{q}_{k-1} + \big(\frac{1}{2} \hat{q}_{k-1} \otimes \mathbf{p}\{w^\text{gyr}_k\}  - \beta \frac{\frac{1}{2} \text{Jac}(f(\hat{q}_{k-1}))^T f(\hat{q}_{k-1})} {\| \frac{1}{2} \text{Jac} (f(\hat{q}_{k-1}))^T f(\hat{q}_{k-1}) \|}\big) \Delta t \\
\end{equation}
Where $\text{Jac}(f(\hat{q}_{k-1}))$ is the Jacobian matrix of the cost-function evaluated at $\hat{q}_{k-1}$ and $\beta$ the optimization parameter.
Although this filter can be used for 6-DOF IMU data, it is typically used on 9-DOF IMUs that also have a tri-axial magnetometer.
The output quaternions of both the Madgwick and Mahony filters will be transformed to Euler angles for comparison with the other methods. 

\subsubsection{The Extended Kalman filter}
The Extended Kalman Filter (EKF) is one of the most-used fusion algorithms for non-linear systems and can also be applied to attitude estimation with a 6-DOF IMU~\cite{Sabatini2011}. 
It uses a model of the system to predict the state at the next time-step and corrects this prediction with measurements using a prediction of the estimate covariance.
This implementation uses the angular rate measured by the gyroscope in the prediction step and the angle measured from the accelerometer measurements in the correction step. 
Since noise on the dynamic model and sensor measurements is defined by covariance matrices, the balance between the gyroscope and accelerometer can be optimized by changing these matrices. 

\subsubsection{A non-neuromorphic neural network}
It is not common to use ANNs for attitude estimation, although there is an increasing interest in using ANNs for low-level estimation and control tasks.
Here, we also compare our neuromorphic algorithm with a more traditional neural network.
As a comparison to the Att-SNN, it was decided to implement a recurrent neural network consisting of two layers with GRUs similar to the one proposed in~\cite{Weber2020}. However, in our implementation the network has Euler-angles in radians as outputs, instead of quaternions to keep it comparable to our Att-SNN, during training and evaluation. 
The recurrency and memory in the GRUs allow the network to store and keep track of the states in the network. This is something a regular feedforward ANN would not be able to do, but that is important for integrating and filtering information over time for state estimation.
The network has the same number of neurons per layer as the Att-SNN and is trained using the same optimizer and training strategy. 
The differences lie in (i) the Att-ANN uses GRUs, while the Att-SNN does not, (ii) the neural dynamics, where the Att-ANN sets its state directly based on all feedforward and recurrent connections, and Att-SNN implements the LIF neuron model.

\section{Results}

In the following, we investigate the performance of the proposed Att-SNN. We first compare the accuracy of the Att-SNN with state-of-the-art, non-neuromorphic attitude estimation filters commonly in control of quadrotors. The spike activity of the Att-SNN is then evaluated with respect to the maneuvers performed by the quadrotor to better characterize the dynamic response of the Att-SNN. Lastly, the neuromorphic state estimator is implemented within the control loop to demonstrate the stability and robustness over time while unknown disturbances are applied to the MAV. 




\subsection{Performance analysis}
\label{subsec:perf_analysis}
\begin{figure}[t]
    \centering
    \includegraphics[width=1\linewidth]{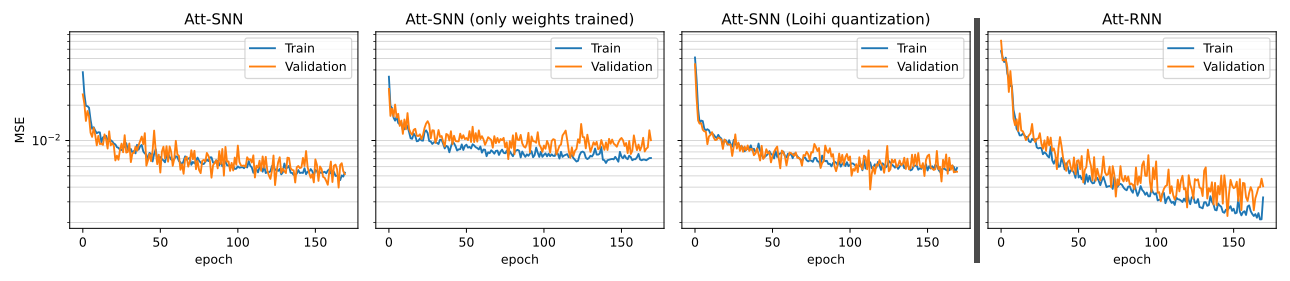}
    \caption{Training loss curve of the Att-SNN vs. the similarly trained Att-RNN with gated recurrent units (GRUs) on training and validation data.}
    \label{fig:loss_curve}
\end{figure}

First, we look at the performance of the neural network methods during training, where we compare three variants of the Att-SNN with the Att-RNN (Fig.~\ref{fig:loss_curve}).
The Att-SNN and the Att-RNN models differ in two aspects. The training of the Att-RNN model results in a more important decrease of the loss function (MSE, Eq.~\ref{eq:mse}) than for the neuromorphic model. Moreover, the validation loss of the Att-SNN model reveals a higher stochasticity (standard deviation of loss minus the moving average of $1.7\times10^{-3}$ vs. $0.9\times10^{-3}$) than that of the Att-RNN model, which can be explained by the non-differentiability of the Heaviside function used in the LIF model (Eq.~\ref{eq:heaviside}). Both networks suffer to some extent from overfitting, since the loss on the training data is lower than that on validation data. By using data with a wide range of dynamic motions and early-stopping during training, generalization can be improved.
The influence of the dataset on the ability of the networks to generalize over data samples never seen during training can be identified by performing a k-fold test. We have split up all PX4 and simulation datasets in 70$\%$ train, 20$\%$ validation and 10$\%$ test data in 5 different folds. We have also included two folds where the network trained on simulation is tested on the PX4 data and vice versa. These are depicted as \textit{sim} and \textit{PX4}.
The mean error and standard deviation (SD) (Table~\ref{tab:cross_val}) for these folds on the corresponding test sets show that the model is resilient to the motions observed in the training data, and new motions in the test set. The error on the test dataset for the second fold is even lower than on the training sets. This can be caused by less dynamic maneuvers in the datasets  that were kept apart for this specific fold. It also shows that training on a single-source dataset results in extreme generalization errors, indicating the importance of mixed-data.

\begin{table}[t]
\centering
\caption{Results of the k-fold cross-validation showing the mean and standard deviation on the test- and training datasets.}
\label{tab:cross_val}
\begin{tabular}{lcccc}
  & \multicolumn{2}{c}{\textbf{Train error}} & \multicolumn{2}{c}{\textbf{Test error}} \\ \cline{2-5} 
Fold  & Mean & SD & Mean & SD  \\ \hline
\textit{1} & $2.09$ & $0.32$ & $3.43$ & $0.44$ \\ \hline
\textit{2} & $2.21$ & $0.50$ & $1.78$ & $0.32$ \\ \hline
\textit{3} & $1.82$ & $0.25$ & $2.42$ & $0.39$ \\ \hline
\textit{4} & $2.05$ & $0.38$ & $2.65$ & $0.48$ \\ \hline
\textit{5} & $2.18$ & $0.51$ & $2.99$ & $0.47$  \\ \hline
\textit{sim} & $1.97$ & $0.23$ & $4.86$ & $0.80$ \\ \hline
\textit{PX4} & $2.20$ & $0.41$ & $5.40$ & $0.65$ \\ \hline
\end{tabular}
\end{table}


To allow the comparison of performance between all filters, the average angle error between the filter and the ground truth was determined for 40 sequences spanning 50 seconds of real-world flight. 



\begin{figure} [t]
     \centering
     \begin{subfigure}[b]{0.48\textwidth}
             \centering
            \includegraphics[width=\textwidth]{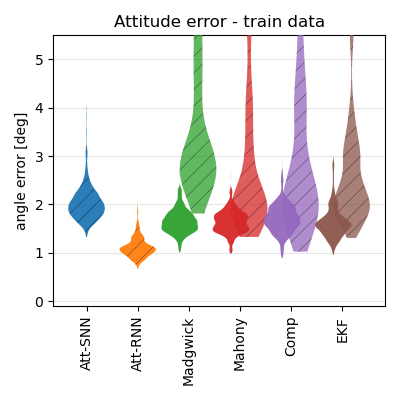}
            \caption{Attitude error on real-world PX4 training dataset.}
            \label{fig:violin_comp_px4_train}
     \end{subfigure}
     \hfill
     \begin{subfigure}[b]{0.48\textwidth}
         \centering
        \includegraphics[width=\textwidth]{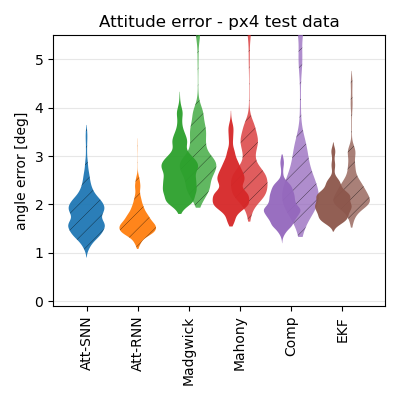}
        \caption{Attitude error on real-world PX4 test dataset.}
        \label{fig:violin_comp_px4_val}
     \end{subfigure}
        \caption{Error distribution of the Att-SNN versus other filters for data obtained in flight. The diagonally hatched plots show the error when the initial angle is not known. Both the Att-SNN and Att-RNN always start without knowledge of the initial angle. Results are given in degrees and combine the results for both the pitch and roll angles.}
        \label{fig:px4_data}
\end{figure}

\begin{figure}
    \centering
    \includegraphics[width=1.1\linewidth]{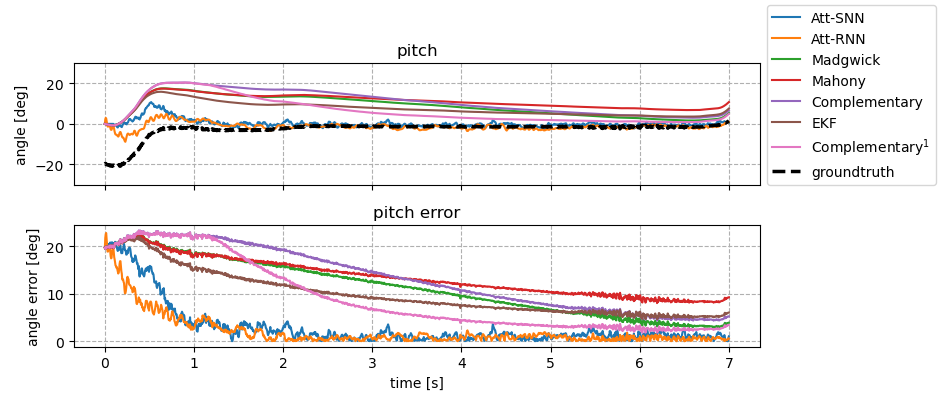}
    \caption{Pitch response and error for different filters w.r.t. ground truth for initial offset of $\approx$20\textdegree. In this case, all conventional filters are tested without any information about the initial angle. Real world PX4 data was used for this comparison. The Complementary$^1$ shows the result of the complementary filter with an adaptation law to allow for quicker convergence.}
    \label{fig:initial_offset}
\end{figure}

In Figure~\ref{fig:px4_data}, we show the overall results of the two neural-based models (Att-SNN and Att-RNN) along with the four filters (Madgwick, Mahony, Complementary, and EKF) with respect to the training PX4 datasets and the validation PX4 datasets. Overall quantitative results are further detailed in Table~\ref{tab:results}. The performance of the standard filters has been measured both with and without prior knowledge of the initial state. The error distribution of the Att-SNN model shows that the network is able to estimate the attitude of the quadrotor with an error comparable to the common filters. The neural-based methods perform better than the standard filters on the training set, when those filters do not have access to the initial state of the quadrotor (i.e., $^1$ in the table). On the test set, all four filters perform slightly worse as compared to the results obtained with the training set, showing that test dataset poses certain challenges that are more difficult for the traditional filters to handle. As expected, the performance of the neural based solutions deteriorates slightly as they now face unknown data. This overfitting may be caused by unknown motions or noise characteristics in the test set and could be reduced by increasing the training data. Adding data from different quadrotors that have different dynamic properties and IMUs with different noise characteristics will increase the generalization properties of the network. However, the average error still remains stable, below $2.5^\circ$ which is in the same range as the errors of the common filters. In Figure~\ref{fig:initial_offset}, we further demonstrate the impact of the initial attitude on the response of the filters. Due to the symmetry of the quadrotor along the $x$ and $y$ axes, only the pitch results are presented. Whereas both the Att-SNN and Att-RNN are able to converge to the real angle in less than a second (200 samples), the three conventional filters have not fully converged after 7 seconds. 
Since the filters all balance between the accuracy on short time-scales given by the gyroscopes and the long time-scale reference of the accelerometer, a large offset at $t=0$ can result in a long convergence. 
This balance is forged by the optimization parameters that were found with the PSO, which remain equal over the course of a sequence. 
These results suggest that the neural networks have learned a way to perform this balance in an adaptive way, trusting the angle $\theta^\text{acc}_k$ more when they largely contradict the current $\hat{\theta}_k$, but rely less on these estimates when the assumption of weak-acceleration does not hold. An implementation of such an adaptive mechanism for attitude estimation with a 9-dof IMU with an adaptive EKF was shown in~\cite{Johansen2017}. To support this hypothesis, the plain complementary filter is adjusted by an adaptive law, trusting more on $\theta^\text{acc}_k$ if the gyroscopes show low angular velocity. If $\lVert \omega^\text{gyr}_k \rVert < 0.1$, the complementary gain $\gamma$ is increased by the difference between the estimated angle $\hat{\theta}_k$ and the angle $\theta^\text{acc}_k$ times a gain $k_a$ chosen as 0.01. 
The result, as seen in Fig.~\ref{fig:initial_offset}, is that the complementary filter recovers faster from an initial offset. Although this does improve the convergence, it is still slower than the neural-based solutions, and is less responsive during aggressive maneuvers. However, it might give some insight in the way the neural-based solutions handle this.

\begin{table}[t]
\centering
\caption{Overall results obtained for all attitude estimation models, compared by means of median, mean and standard deviation on training and test data. Standard filters include results for training and test with and without prior knowledge on the initial state. Results are expressed in degrees.}
\label{tab:results}
\begin{threeparttable}
\begin{tabular}{lcccccc}
\cline{2-7}
 & \multicolumn{3}{c}{\textbf{Training set}} & \multicolumn{3}{c}{\textbf{Test set}} \\ \cline{2-7} 
 & Median & Mean & SD & Median & Mean & SD \\ \hline
\textit{Att-SNN}\tnote{1} & $1.90$ & $1.56$ & $0.24$ & $2.79$ & $2.79$ & $0.37$ \\ \hline
\textit{Att-RNN}\tnote{1} & $\mathbf{1.53}$ & $1.91$ & $0.28$ & $2.45$ & $2.45$ & $0.35$ \\ \hline
\textit{Madgwick} & $2.79$ & $2.81$ & $0.36$ & $2.89$ & $2.92$ & $0.41$ \\
\textit{Madgwick}\tnote{1} & $3.03$ & $3.22$ & $0.83$ & $3.06$ & $3.34$ & $1.11$ \\ \hline
\textit{Mahony} & $2.56$ & $2.57$ & $0.36$ & $2.53$ & $2.54$ & $0.33$ \\
\textit{Mahony}\tnote{1} & $2.69$ & $2.85$ & $0.74$ & $2.59$ & $2.84$ & $0.89$ \\ \hline
\textit{Complementary} & $2.42$ & $2.46$ & $0.42$ & $\mathbf{2.13}$ & $2.13$ & $0.35$ \\
\textit{Complementary}\tnote{1} & $2.60$ & $2.98$ & $1.24$ & $2.36$ & $2.71$ & $1.36$ \\ \hline
\textit{EKF} & $2.67$ & $2.75$ & $0.52$ & $2.36$ & $2.40$ & $0.44$ \\
\textit{EKF}\tnote{1} & $2.65$ & $2.71$ & $0.51$ & $2.49$ & $2.64$ & $1.00$ \\ \hline
\end{tabular}
\begin{tablenotes}
    \item[1] No knowledge of initial state.
\end{tablenotes}
\end{threeparttable}
\end{table}


\subsection{Spiking activity}
 
\begin{figure} [t]
     \centering
     \begin{subfigure}[t]{0.48\textwidth}
         \centering
        \includegraphics[width=\textwidth]{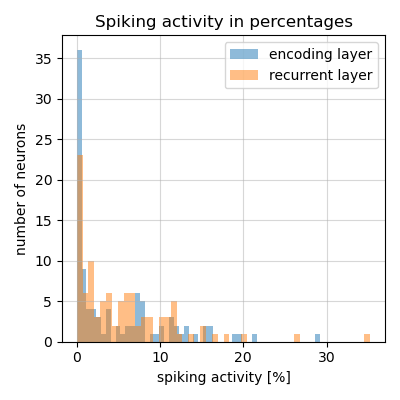}
        \caption{Histogram of the spiking activity in percentages of all neurons, averaged over the datasets}
        \label{fig:spikes_per_neuron}
     \end{subfigure}
     \hfill
     \begin{subfigure}[t]{0.48\textwidth}
             \centering
            \includegraphics[width=\textwidth]{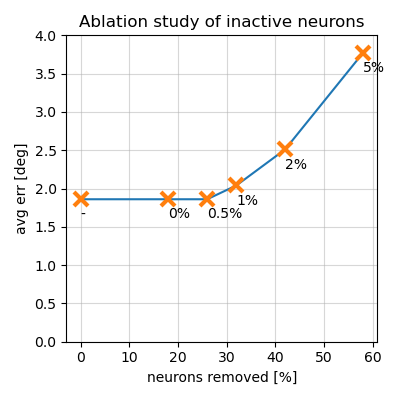}
            \caption{Effect of removing sparsely spiking neurons on accuracy to reduce the overall network-size and prospective energy consumption in an end-to-end solution.}
            \label{fig:ablation}
     \end{subfigure}
        \caption{Investigation in the spiking activity of all neurons in the Att-SNN and the effect of removing sparsely spiking neurons.}
        \label{fig:spiking_activity}
\end{figure}

After training the Att-SNN model, we measured the overall spiking activity over the datasets of both the encoding and the hidden layers to evaluate the sparsity of the network. A histogram of the average spiking activity per neuron is given in Figure~\ref{fig:spikes_per_neuron}. Results show that, out of all neurons, $27$\% of neurons in the encoding layer and $10$\% of neurons in the recurrent layer do not spike at all, and that an extra $7$\% and $8$\% respectively spike less than $0.5$\% of the time. To understand if these sparsely spiking neurons have a significant effect, an ablation study has been performed. By pruning all neurons that spike less than $\mathbf{x}$\% of the time for a subset of data, the effect on the accuracy for the rest of the data can be established. The results are shown in Figure~\ref{fig:ablation}. Pruning the neurons that do not exceed an average spiking activity of $0.5\%$ did not modify the performance of the network. This corresponds to a total of approximately $25\%$ of the neurons of the Att-SNN model, which amounts to a reduction of almost half of the synapses while making the network more efficient.

\subsection{Input dataset manipulation}

\begin{figure}[t]
    \includegraphics[width=1\textwidth]{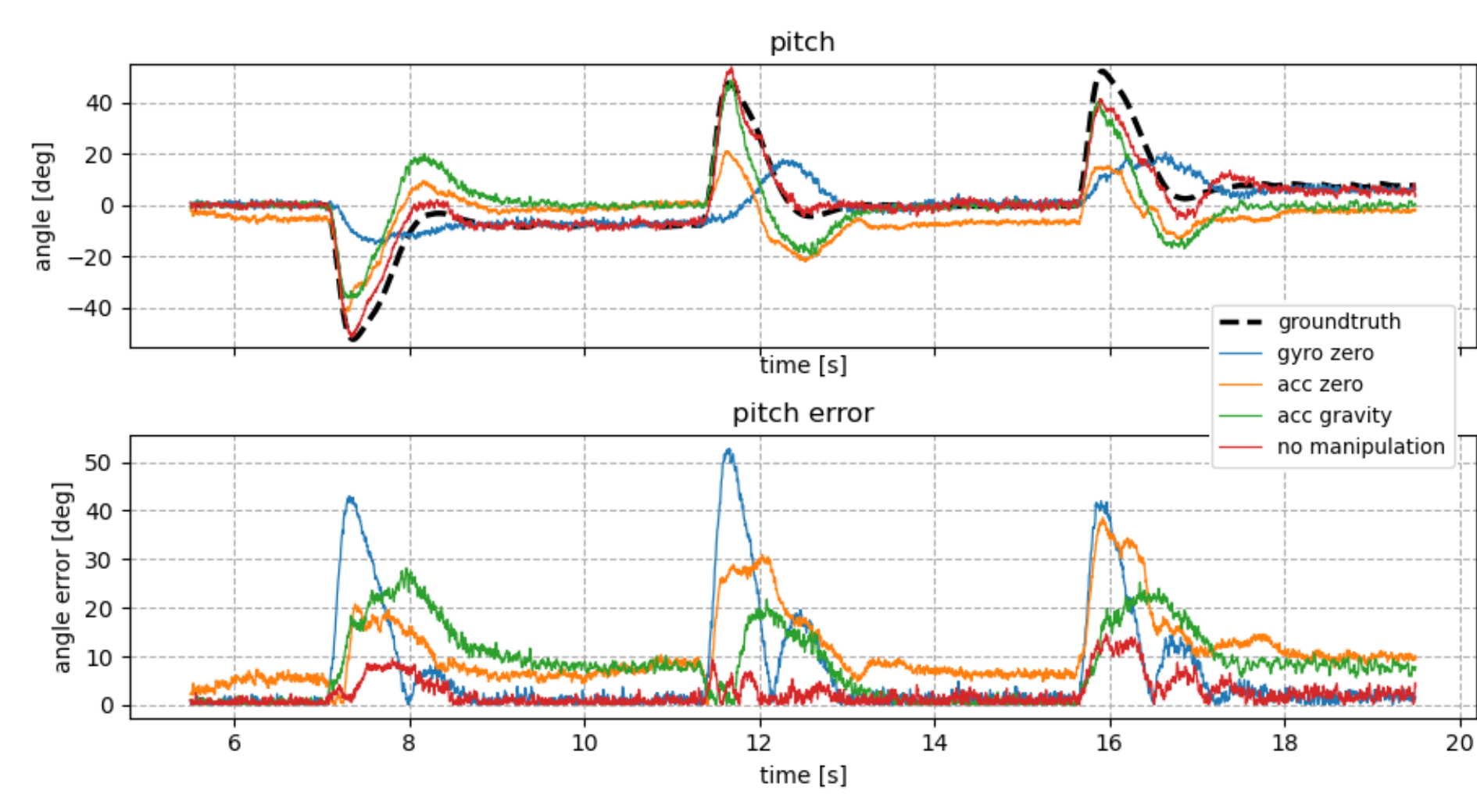}
    \caption{Response of the SNN for different types of input manipulation. Both the pitch estimation and error of the network is shown for (i) no data manipulation, (ii) all gyroscope values zero, (iii) all accelerometer values zero, and (iv) all accelerometer values pointing towards gravity in world-frame ($\theta^\text{acc}_k = 0$).}
    \label{fig:input_manipulation}
\end{figure}

In order to obtain some preliminary understanding of the functioning of the SNN, we have evaluated the fusion of sensory information by manipulating sensory inputs.
In particular, we compared the output of the Att-SNN model during a constant velocity flight where the attitude is kept constant. In Figure~\ref{fig:input_manipulation}, the response of the network is provided for the pitch angle of the quadrotor obtained in simulation and in the following conditions: (i) the gyroscope is removed, (ii) the accelerometer is removed, (iii) the accelerometer data all point towards gravity in the world frame (resulting in $\theta^\text{acc}_k=0$), and (iv) no manipulation (i.e., the initial Att-SNN). We observe that removing the angular velocity information provided by the gyroscope results in large errors during fast motion, but remains correct when the motion follows a constant angle. This shows that the Att-SNN model uses the accelerometer data for a long time scale estimation of the angle, similar to how the common filters use it. When the accelerometer data is replaced by the gravity vector in the world frame, effectively measuring a zero angle at all times, this results in an estimate that is quite reliable during fast maneuvers.
However, during constant velocity flight the output of the estimator is approximately zero, since it has no absolute measurement of the real angle. 
This shows that the network uses the accelerometer as an absolute reference to the global attitude, just like the comparison filters do.


\subsection{Energy consumption and evaluation frequency on neuromorphic hardware} 
\begin{figure} [t]
     \centering
     \begin{subfigure}[b]{0.32\textwidth}
             \centering
            \includegraphics[width=\textwidth]{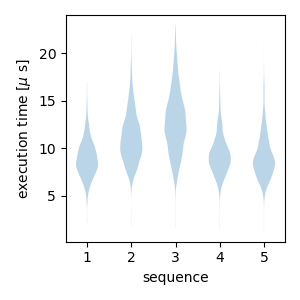}
            \caption{Execution time for 5 sequences of 2000 time steps}
            \label{fig:loihi_1}
     \end{subfigure}
     \hfill
     \begin{subfigure}[b]{0.32\textwidth}
         \centering
        \includegraphics[width=\textwidth]{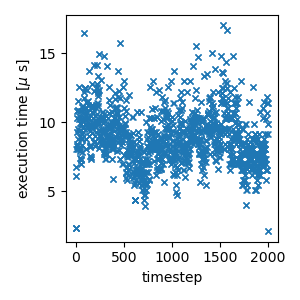}
        \caption{Execution times for an example sequence}
        \label{fig:loihi_2}
     \end{subfigure}
     \hfill
     \begin{subfigure}[b]{0.32\textwidth}
         \centering
        \includegraphics[width=\textwidth]{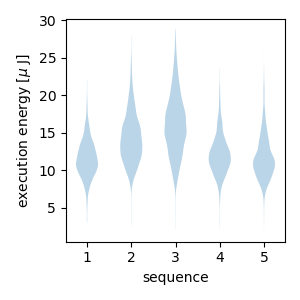}
        \caption{Energy consumption for 5 sequences of 2000 time steps}
        \label{fig:loihi_3}
     \end{subfigure}
    \caption{Energy and evaluation frequency of the Att-SNN on the Nahuku board that has 32 Loihi chips.}
    \label{fig:loihi_eval}
\end{figure}

To evaluate the potential of deploying these networks on neuromorphic processors, the update frequency and energy consumption of the network were examined.
Since gathering energy benchmarks on the Kapoho Bay, used on our quadrotor, is not possible, the results shown in Figure~\ref{fig:loihi_eval} are obtained with a Nahuku board that has 32 Loihi chips by utilizing the energy and execution time probes as provided in the NxSDK. 
The average execution time of the spiking recurrent layer on hardware was only $10\mu$s, which corresponds to 100~000 Hz. As comparison, the execution time of the layer was also evaluated using a PyTorch implementation on a laptop with a Nvidia GTX 1650Ti and a 12-core Intel i7-10750H CPU. The average execution time for a single timestep of the Att-SNN here was $555\mu$s, $\approx55$ times as high as running the network on the neuromorphic processor. Also the execution time of the Att-RNN was measured in the same way, and the average execution time here was $150\mu$s, still an order of magnitude higher than on the neuromorphic processor. 
Also the number of Floating-point Operations (FLOPS) per method per timestep have been calculatd. The Att-SNN requires only ($M^2 + MN + 2M$) FLOPS, with $M$ the number of neurons in the hidden layer, and $N$ the amount of inputs to this layer. The Att-RNN with gated-recurrent units requires approximately ($3M^2 + 3MN + 3M$) FLOPS for a single hidden layer, which is almost 3 times more. 
The average energy consumed per time step was $13.2\mu$J.
In our hardware architecture with the PX4 mini flight-computer connected to an UP Squared computer using Fast-RTPS, we were able to run it at the frequency of incoming IMU data, 200Hz. It should be noted this looprate is dependent on the data transferring in the Von Neumann processors. In an end-to-end neuromorphic pipeline, this could be orders of magnitude faster. Since we run the Loihi at this frequency, the latency from input sample to estimate is governed by the delays in the network. In our implementation, this means an estimate is produced after three timesteps, resulting in a latency of 15ms. 


\section{Discussion}
This work demonstrates that a small recurrent spiking neural network is able to estimate the attitude of a quadrotor in flight. 
In terms of accuracy, it is competitive with filters commonly used in flight-controllers, such as the Mahony-filter. 
Our approach is not data hungry, as common AI algorithms are, and only requires data obtained in simulation (easily accessible, and available open-source) supplemented with short sequences obtained with a real quadrotor (for our training, less than 20 minutes of real data was used). 
Our experiments have shown that little data already suffices for successful attitude estimation in nominal flight conditions.
However, future work could consider larger datasets with a larger variety of flight maneuvers, and naturally a larger range of sensor biases.
This would help to find out how well trained networks generalize to unseen conditions.

These results are obtained with focus on achieving a small-scale network. 
By increasing the size of the network, the training performance can be increased but the network might suffer more from overfitting. Already in our limited sized network, it was visible that training allowed for overfitting on the training data. This effect might be limited by supplying the training with more samples from different real world quadrotors or making the network more resilient to biases in the input (such as constant offsets of the measured accelerations) by adding parametrization. This parametrization could be obtained by adding adaptivity to the synaptic connections, for which a suitable online learning rule could be designed. 
Besides, minimizing size is in line with our goals of obtaining miniature-scale robots, capable of performing autonomous missions. 
Increasing the size too much will also constrain the learning, due to an increased dimensionality, resulting in higher memory usage and training time.

A main potential criticism on our work could be that the plain complementary filter, achievable with only a couple lines of code in a microcontroller, is still a valuable choice for performing the task of attitude estimation based on IMU-data. 
Executed on a widely available and cheap micro-controller, it estimates states fast and with little energy expenditure. However, we aim for a fully neuromorphic pipeline, so that all processing, from very intensive visual processing to less intensive state estimation and control, can happen on a single neuromorphic chip. Hence, we also need to perform attitude estimation with an SNN. Moreover, from a scientific viewpoint, we are interested in understanding how (spiking) neural networks solve this task, potentially unveiling new strategies or delivering new hypotheses on sensor fusion in flying animals like small insects. Already in this study, the SNN showed the interesting property of converging quicker to the attitude when not initialized at the ground-truth attitude. In future work, we plan to delve deeper into the detailed workings of SNNs estimating attitude.

Currently, the work has the limitation that the Att-SNN only estimates the angles necessary for position control, and excludes yaw.
Synthesizing an unbiased, non-diverging yaw-estimate using angular velocities and linear accelerations exclusively is not possible, so extra sensors would need to be included. 
A common option is the 3-dof magnetometer, measuring the Earth's magnetic field and therefore supplying the estimator with an absolute measurement of the quadrotors heading. 
Although these sensors are very useful in large drones flying at great altitudes in non-urban environments, the readings are heavily influenced by disturbances in the magnetic field. 
these disturbances can be caused by electronic appliances in the vicinity or even by the motors of the quadrotor itself. In future work, adding data from other sensors or the output of a controller will be studied to capture a full 3d model of a quadrotor in flight. This will pose interesting new challenges such as dynamics changing over a flight because of a draining battery that might be addressed by an SNN that features adaptivity. 

The frequency of the network in this study was chosen to be 200Hz, since this matched the output of the sensor data from the PX4 flight-controller. 
In operation, the input-data streamed to the network has to be of the same rate as during training, because of the time-constants that characterize the neuron dynamics. 
During this research, it was found that training on even higher data rates, further reduced the estimation-error. 
Together with the promises of extremely high update rates of neuromorphic hardware, this is encouraging for further research.

Considering the recurrent layer encodes the attitude (or rate of change of the attitude) necessary for control, the decoding layer can be replaced by another spiking layer that can be trained to perform control tasks, either by supervised learning mimicking a baseline controller, or it can be trained in a RL framework, as is discussed in \cite{Hwangbo2017}.
These are the next steps that follow naturally from our research.

\section{Conclusion}
In this paper, we have presented an SNN model, called Att-SNN, that can be employed as an attitude estimator in an end-to-end neuromorphic control pipeline. This implementation builds upon three contributions. First, we have shown that the Att-SNN can perform state estimation tasks in highly dynamic systems such as MAVs, with competitive performance when compared to state-of-the-art non-neuromorphic methods and conventional recurrent ANNs. Second, we successfully implemented the Att-SNN on the Loihi neuromorphic processor, showing outstanding energy and time efficiency, therefore paving the way towards fully embedded neuromorphic control onboard MAVs. Third, this study shows for the first time that these networks can be used as an estimator in the control-loop, showing that small errors do not accumulate over time. Furthermore, our work shows an efficient method of encoding floating point sensor data into binary spikes without having to study the tuning curves of rate-coded neurons or finding the optimal distribution of neurons in population coding. Several prospective future research directions are recognized. This includes extending the estimation with neuromorphic control and reducing the effect of uncertainties due to sensor noise by employing online adaptivity. All together, the Att-SNN neuromorphic attitude estimation model will help closing the neuro-biologically-inspired control loop from sensor to actuator in critical embedded systems such as MAVs. 


\section*{References}
\bibliographystyle{IEEEtran}
\bibliography{biblio}

\end{document}